# ENHANCING ROAD SIGNS SEGMENTATION USING PHOTOMETRIC INVARIANTS

*Tarik Ayaou[1], Azeddine Beghdadi[2], Karim Afdel[3], Abdellah Amghar[1]*

[1]LMTI, Faculty of Sciences, IbnouZohr University Agadir, Morocco
Email: tarik.ayaou@edu.uiz.ac.ma
[2]L2TI, Université Paris 13, Sorbonne Paris Cité, Villetaneuse, France
[3]LabSiv, Faculty of Sciences, IbnouZohr University, Agadir, Morocco

## ABSTRACT

Road signs detection and recognition in natural scenes is one of the most important tasksin the design of Intelligent Transport Systems (ITS). However, illumination changes remain a major problem. In this paper, an efficient approach of road signs segmentation based on photometric invariants is proposed. This method is based on color information using a hybrid distance, by exploiting the chromatic distance and the red and blue ratio, on l$\theta\Phi$ color space which is invariant to highlight, shading and shadow changes. A comparative study is performed to demonstrate the robustness of this approach over the most frequently used methods for road sign segmentation. The experimental results and the detailed analysis show the high performance of the algorithm described in this paper.

## 1. INTRODUCTION

Road and traffic sign detection and recognition represent one of the principal issues of any Intelligent Transportation System (ITS) [1]-[8].This is due to the importance of the road signs to the road users. They give them important information by using unique shapes and colors distinguishing them from the other objects. These properties were exploited in different approaches to detect and recognize road signs. Generally, a road sign detection and recognition system is composed of three stages;segmentation, detection and recognition. The first stage is very useful to reduce the space of research by using color information [2][10]. The second one aims to find the objects from the segmented Region of Interest (ROI) which can be considered as probable road signs [1][3]. In the last stage, the detected objects are also processed and fed into a system to identify their types in order to determine false and positives detection [4][6].

The detection of the road signs from outdoor images is the most complex step in the automatic traffic sign detection and recognition system [11]. Many interesting works focus on the recognition stage [9]-[11], although the segmentation and detection steps are the most important.

However, image segmentation and road sign detection are quite difficult when dealing with traffic road scenes. This is mainly due to some physical factors, such as weather, illumination change and material reflectance properties that affect the quality of the acquisition. Indeed, the illumination changes represent the most important difficulty in image segmentation and in many computer vision applications such as driver assistance systems. Such illumination changes include for instance shadow/shading, specular reflections and globally varying illumination [13] [14]. Some works, try to tackle these problems by using optical image acquisition models based on color space representations [23]. The acquired image is considered as a physical signal containing the reflectance, illumination and shape of objects observed in the scene. However, the information that could be extracted from the signal by using these models is affected by uncontrolled factors related to acquisition conditions. A lot of works has been done to extract scene informationin a way that is invariant to illumination and reflectance changes[1] [2][3].

For image segmentation, the problem of illumination changes could be overcome by using some color spaces such as the normalized RGB [1] or Hue-Saturation-Value (HSV) [2] [13]. Furthermore the spherical interpretation of the rgb color space shows good results [14].Those methods are based on dichromatic model of reflectance which is independent of diffuse shading and depends only on the spectral reflectance of the surface and the spectral power distribution (SPD) of the illumination [16].

In this paper we propose a new method based on photometric invariants derived from DMR expressed in a color spherical space. This method is compared to some methods of the state-of-the-art. The rest of the paper is organized as follows. Section 2 provides a brief overview of color image formation models and photometric invariants used in computer vision. Section 3 contains the description of the proposed method. The results and discussion are given in section 4. The paper ends with the conclusion and possible future research.

## 2. BACKGROUND: COLOR AND PHOTOMETRIC INVARIANTS

Many color models and color features to be used for the purpose of image segmentation, recognition and classification have been proposed in the literature to solve the problem of changes of some physical properties during image acquisition [2][3]. It's worth nothing that the reflectance is one of the most relevant physical properties in color image processing and analysis. In this paper we concentrate on the dichromatic reflection model (DRM) which is suitable for road sign image. Indeed, Tominaga demonstrated that the paint is a dielectric material that could be well described by DRM [7]. Therefore it is appropriate for analyzing images of road signs that are characterized by colorful paints.

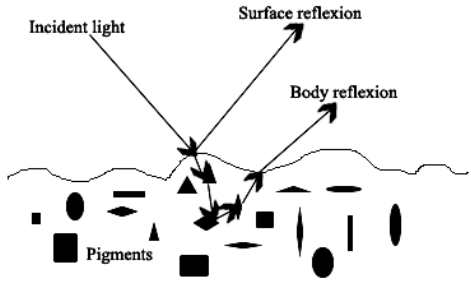

**Fig. 1.** Interaction of electromagnetic power with a dielectric material.

The reflectance signal, as the ratio between the reflected electromagnetic power and the incident one, especially the specular reflectance is the most difficult problem in computer vision. The origin of this problem is that the specular reflectance produces a single reflected ray, while the quality of reflection depends on the quality of the reflecting interface. This interface tends to become perfectly reflecting when the size of the defects of the interface is less than or of the order of the incident wavelength (red: 800 nm, blue: 460nm), otherwise the interface tends to distort the signal reflected.

In order to solve the problem related to photometric variants, the common and known approach is to explore functions that are invariants to photometric changes. A photometric invariant is the representation of a pixel color $c(r, g, b)$ as a quantity independent of the intensity of the pixel, the reflecting interface properties or the camera characteristics. The transformation of the $rgb$ color space to the spherical coordinates $(l,\theta,\phi)$ is used. Where, both angles $\theta$ and $\phi$ (Equation (3) and (4)) are invariant with respect to highlight, shadow and shading. However, the color magnitude $l$ is not a photometric invariant.

The transformation of the pixel color components from the $rgb$ color space into the spherical one $(l,\theta,\phi)$ is expressed as follows:

$$\begin{cases} l = \sqrt{r^2 + g^2 + b^2} \\ \phi = \arctan\left(\frac{g}{r}\right) \\ \theta = \arccos\left(\frac{\sqrt{r^2 + g^2}}{\sqrt{r^2 + g^2 + b^2}}\right) \end{cases} \quad (1)$$

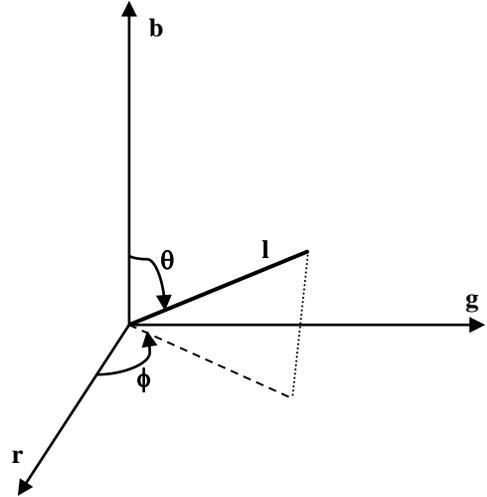

**Fig. 2.** Presentation of rgbcolor space in spherical coordinate $l\theta\Phi$

Where $l$ denotes, the magnitude of the color vector and $\theta$ and $\varphi$ are two angles that represent, respectively, the longitude and the latitude.

The rgbcolor pixel, acquired from a Lambertian surface, is given according to the model introduced in [20]:

$$c_k = \sigma I k_1 \lambda_k^{-5} e^{\frac{-k_2}{T\lambda_k}} S(\lambda_k) q_k \quad (2)$$

Where $\sigma$ is the Lambertian shading, $I$ is the overall light intensity, $k_1$ and $k_2$ are two constants, $S(\lambda_k)$ is the surface spectral reflectance function and $q_k$ represents the camera sensor sensitivity and $k=1...3$ is the index of the $rgb$ color channels. Temperature T characterizes the lighting color.

Using equations (1) and (2), $\theta$ and $\phi$ are expressed as follows:

$$\theta = \tan^{-1}\left(\frac{g}{r}\right) = \tan^{-1}\left(\frac{\lambda_2^{-5} e^{\frac{-k_2}{T\lambda_2}} S(\lambda_2) q_2}{\lambda_1^{-5} e^{\frac{-k_2}{T\lambda_1}} S(\lambda_1) q_1}\right) \quad (3)$$

$$\phi = \cos^{-1}\left(\frac{\sqrt{r^2+g^2}}{\sqrt{r^2+g^2+b^2}}\right)$$

$$= \tan^{-1}\left(\frac{\sqrt{A+B}}{\sqrt{M}}\right) \quad (4)$$

with

$$A = \left(\lambda_1^{-5} e^{\frac{-k_2}{T\lambda_1}} S(\lambda_1) q_1\right)^2 \quad (5)$$

$$B = \left(\lambda_2^{-5} e^{\frac{-k_2}{T\lambda_2}} S(\lambda_2) q_2\right)^2 \quad (6)$$

and

$$M = \left[A + B + \left(\lambda_3^{-5} e^{\frac{-k_3}{T\lambda_3}} S(\lambda_3) q_3\right)^2\right] \quad (7)$$

## 3. PROPOSED METHOD

To ensure a robust segmentation, it is very important to make it invariant under illumination changes. As explained above, the spherical representation of color is invariant to shading and highlight. This property is then exploited in the design of the proposed segmentation approach. This method is based on a hybrid distance $Hdi$ defined as:

$$Hdi = \alpha_i(x) * Cdi(p, o) \quad (8)$$

where

$$Cdi(p, o) = \sqrt{(\theta_o - \theta_p)^2 + (\phi_o - \phi_p)^2} \quad (9)$$

$$\alpha_r(x) = \begin{cases} 0 & x \le a \\ x & a < x < b \\ 1 & x \ge b \end{cases} \quad (10)$$

and

$$\alpha_b(x) = \begin{cases} a-x & , x \le a \\ 1 & , x \ge a \end{cases} \quad (11)$$

Where $Cdi$ is the chromatic distance corresponding to the Azimuth and Zenith, between a given pixel $p$ and its associated value $o$ obtained from a training set of pixels.

The parameters $a$ and $b$ are obtained from a training set from different road signs images representing various lighting conditions.

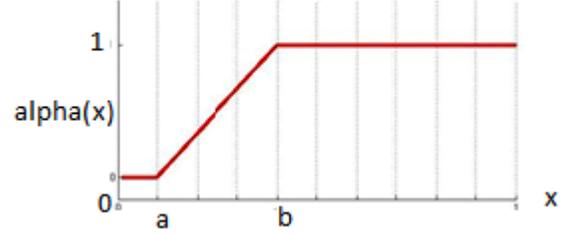

**Fig.3.** Transfer function of the red extraction.

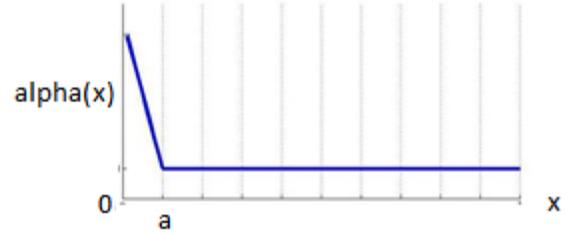

**Fig.4.** Transfer function of the blue extraction.

Finally, a thresholding process is performed to classify pixels as defined below.

$$\begin{cases} Hdr > r/b, & \text{The pixel is red} \\ Hdb < r/b, & \text{The pixel is blue} \end{cases} \quad (14)$$

## 4. RESULTS AND DISCUSSION

To evaluate the performance of the proposed method, we apply the connected components approach on the segmented images of the German Traffic Sign Detection Benchmark (GTSDB) dataset and an England Dataset.

The first dataset contains 300 images (1360, 800 pixels) and the second contains 128 image (640, 480 pixels), taken under different illumination conditions. A comparative study is also performed to evaluate the proposed approach against the most frequently ones, i.e., *HSV* color space

[10][2][21], red enhancing approach [4] and Log-Chromatic transformation [22]. In fig.1, the Precision-Recall curve shows that the proposed method outperforms the other ones.

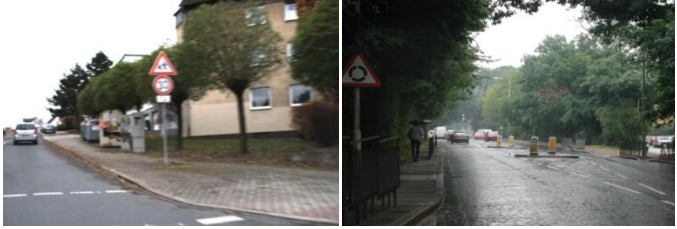

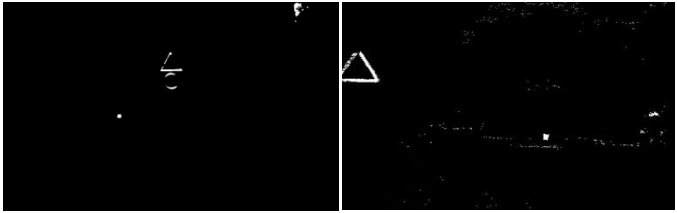

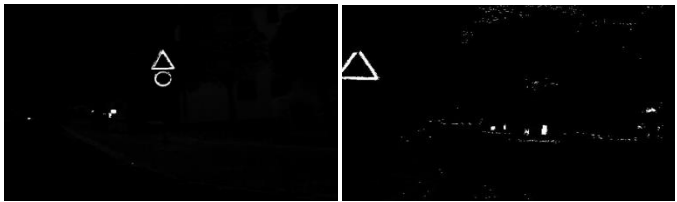

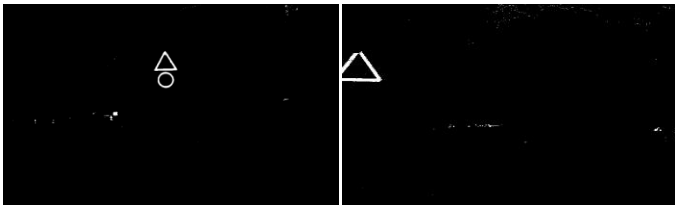

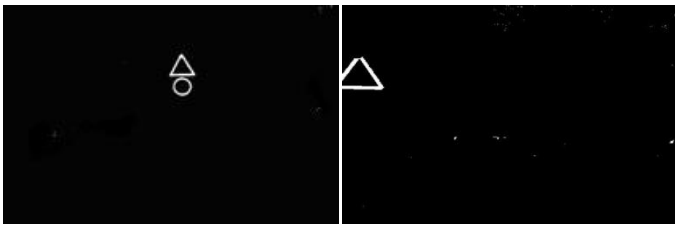

**Fig.2 .**Segmentation results
Top-to-Bottom rows: 1: Original Image, 2: HSV, Red-Enhancing, 3: Log-Chromatic, 4: Our approach

Figure2 illustrates examples of segmented images using the different methods used in this work. From the experimental results it can be seen that the HSV color space and the red-enhancing method perform poorly in comparison to the Log-Chromatic and $l\theta\phi$ based methods.

From the detection rates using the test dataset, summarized in Table1, it can be concluded that $l\theta\phi$ color space is robust for color based segmentation of road signs under varying illumination conditions. An accuracy of 90% can be achieved when using $l\theta\phi$ color space with the Hybrid Distance in Spherical Coordinates (HDSC).

The accuracy rate and the performance of the proposed approach are measured by applying it on the two sets. The performance evaluation is based on the recall and precision values, which are defined as follows:

$$recall = \frac{true\ positive\ detected\ (TP)}{total\ true\ positves} \times 100\% \quad (14)$$

$$precision = \frac{true\ positive\ detected\ (TP)}{all\ detections} \times 100\% \quad (15)$$

|  | GTSDB dataset | | England Dataset | |
|---|---|---|---|---|
| Method | Precision (%) | Recall (%) | Precision (%) | Recall (%) |
| HSV | 70.32 | 53.15 | 78.58 | 69.75 |
| Red-Enhancing | 74.25 | 62.86 | 79.42 | 68.34 |
| Log-Chromatic | 81.12 | 71.56 | 91.86 | 82.71 |
| HDSC | 92.52 | 83.20 | 96.65 | 84.48 |

**Tab.1 .**Overall detection accuracy obtained when using theconnected components approach for the different segmentation methods

## 5. CONCLUSION AND PERSPECTIVES

In this paper a robust and efficient approach of road signs segmentation is proposed and evaluated. The proposed segmentation method is based on the dichromatic reflection model, expressed in spherical coordinates.The obtained results prove that by exploiting the invariance to shading, shadow and highlights of HDSC, expressed in $l\theta\phi$ color space, a significant improvement of the segmentation performance could be achieved. It is worth noticing that the proposed scheme is of low computational complexity and does not require any pre-processing. This approach seems to be suitable for real-time applications. Regarding our future work, we intend to conceive an embedded system for road sign detection and recognition as driver assistant.